\documentclass[runningheads]{llncs}

\usepackage[mobile]{eccv}

\usepackage{eccvabbrv}

\usepackage{graphicx}
\usepackage{booktabs}

\usepackage{amsmath} 
\usepackage{multirow}  
\usepackage{tikz} 
\usepackage{xcolor}  
\definecolor{seen_color}{RGB}{250,132,161}

\usepackage[accsupp]{axessibility}  

\usepackage[pagebackref,breaklinks,colorlinks,citecolor=eccvblue]{hyperref}

\usepackage{orcidlink}

\begin{document}

\title{AlignZeg: Mitigating Objective Misalignment for Zero-shot Semantic Segmentation}

\titlerunning{AlignZeg}

\author{Jiannan Ge\inst{1} \and
Lingxi Xie\inst{2} \and
Hongtao Xie\inst{1} \and 
Pandeng Li\inst{1} \and 
Xiaopeng Zhang\inst{2} \and 
Yongdong Zhang\inst{1} \and 
Qi Tian\inst{2}
}

\authorrunning{J. Ge et al.}

\institute{University of Science and Technology of China \and
Huawei Inc. }

\maketitle

\begin{abstract}
A serious issue that harms the performance of zero-shot visual recognition is named \textbf{objective misalignment}, 
\textit{i.e.}, the learning objective prioritizes improving the recognition accuracy of seen classes rather than unseen classes, while the latter is the true target to pursue.
This issue becomes more significant in zero-shot image segmentation because the stronger (\textit{i.e.}, pixel-level) supervision brings a larger gap between seen and unseen classes.
To mitigate it, we propose a novel architecture named \textbf{AlignZeg}, which embodies a comprehensive improvement of the segmentation pipeline, including proposal extraction, classification, and correction, to better fit the goal of zero-shot segmentation. 
\textbf{(1) Mutually-Refined Proposal Extraction.} AlignZeg harnesses a mutual interaction between mask queries and visual features, facilitating detailed class-agnostic mask proposal extraction. \textbf{(2) Generalization-Enhanced Proposal Classification.} AlignZeg introduces synthetic data and incorporates multiple background prototypes to allocate a more generalizable feature space. 
\textbf{(3) Predictive Bias Correction.} During the inference stage, AlignZeg uses a class indicator to find potential unseen class proposals followed by a prediction postprocess to correct the prediction bias. 
Experiments demonstrate that AlignZeg markedly enhances zero-shot semantic segmentation, as shown by an average 3.8\% increase in hIoU, primarily attributed to a 7.1\% improvement in identifying unseen classes,
and we further validate that the improvement comes from alleviating the objective misalignment issue.

\end{abstract}

\section{Introduction}
\label{sec:intro}

\begin{figure}[t]
  \centering
  \includegraphics[width=1\linewidth]{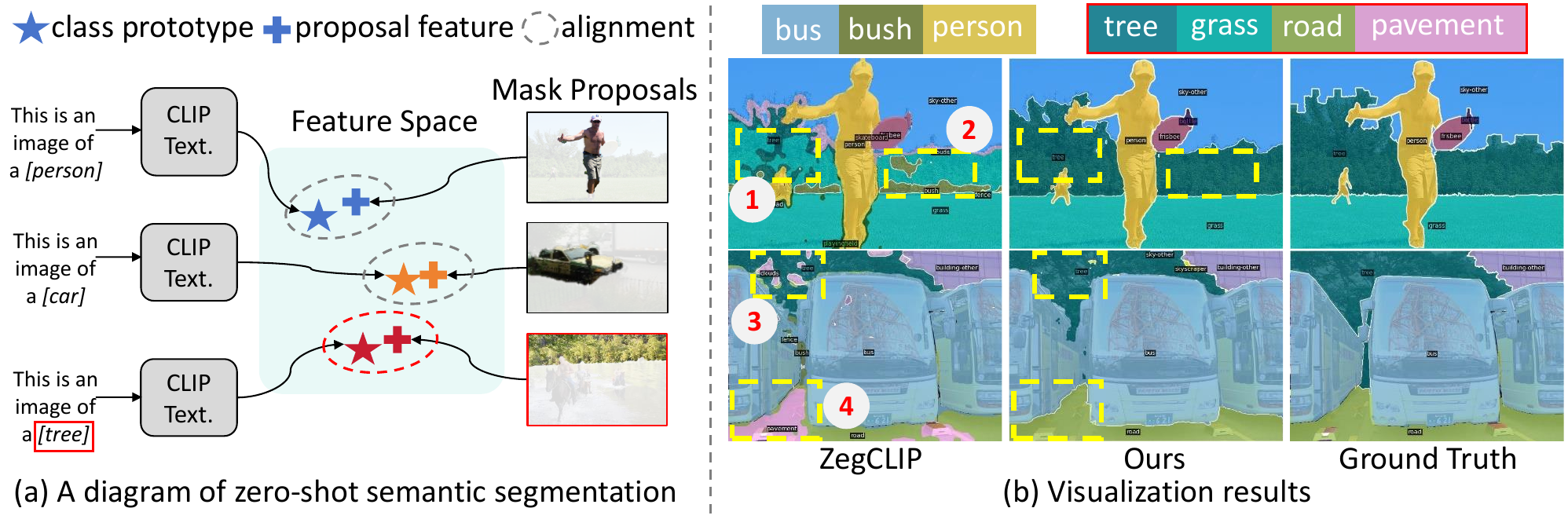}
   \caption{(a) Proposal features are aligned with class prototypes. ``CLIP Text.'' represents CLIP Text Encoder. (b) 
   Red boxes select the unseen classes. 
   Yellow dashed boxes 
   select the misclassified areas, \textit{e.g.},  ``\textit{\textcolor{red}{tree}}'' (unseen) $\rightarrow$ ``\textit{bush}'' (seen) {\large{\textcircled{\small{2}}}\normalsize}, ``\textit{\textcolor{red}{road}}'' (unseen) $\rightarrow$ ``\textit{\textcolor{red}{pavement}}'' (unseen) {\large{\textcircled{\small{4}}}\normalsize}. 
   These errors show the objective misalignment issue.
   }
   \label{fig:intro}
\end{figure}
Semantic segmentation~\cite{pastore2021closer,chen2023exploring,cai2023mixreorg,cho2023cat,luo2023segclip,wu2023diffumask,huo2022domain, kirillov2023segany} is a fundamental task in computer vision that is critical for image understanding. Traditional segmentation \cite{cheng2022masked, liu2022dynamic,jin2021mining} approaches depend on large amounts of data with detailed annotations to perform effectively.
Given the vast and diverse nature of real-world data, it is not feasible to annotate every new image. 
Zero-shot semantic segmentation \cite{bucher2019zero} has emerged to address this challenge, leveraging the model trained on a limited dataset to segment and recognize classes not seen during training. 

Recently, the introduction of large-scale vision-language models, such as CLIP~\cite{radford2021learning}, has advanced this task further. As shown in \cref{fig:intro} (a), the visual features of seen classes are aligned with the semantic features, \textit{i.e.}, class prototypes, which enables the model to detect unseen classes. 
Models such as SimBaseline~\cite{xu2022simple}, Zegformer~\cite{ding2022decoupling}, and OVSeg~\cite{liang2023open} have proposed a two-stage scheme that first generates class-agnostic mask proposals and then extracts corresponding mask visual features for zero-shot classification. 
This scheme effectively extends CLIP's zero-shot abilities to the pixel level. 
However, these methods often necessitate multiple forward passes per image, leading to inefficiency. 
In response, models like ZegCLIP~\cite{zhou2023zegclip}, DeOP~\cite{han2023open}, and SAN~\cite{xu2023side} construct their networks based on the CLIP image encoder for a streamlined process. 
However, these methods often overlook a fundamental issue which is the significant divergence between their objectives, \textit{i.e.}, improving the performance of seen classes, and the principles of zero-shot learning, a phenomenon we refer to as ``\textbf{objective misalignment}''. 
In the semantic segmentation task, characterized by a strong supervision signal, this misalignment issue becomes more pronounced. It not only constrains the full potential of CLIP's zero-shot performance but also introduces biases into the model's predictions. 
As illustrated in \cref{fig:intro} (b), this misalignment tends to cause inaccurate segmentation results on unseen classes. For example, the model might mistakenly segment ``\textit{\textcolor{red}{tree}}'' class as semantically similar categories like ``\textit{\textcolor{red}{grass}}''\large{\textcircled{\small{1}}} \normalsize and ``\textit{bush}''\large{\textcircled{\small{2}}}\normalsize.

To address the issue mentioned above, we propose a novel framework named AlignZeg, which aligns the training and inference objectives of the model with zero-shot tasks. Initially, 
we introduce the \textbf{Mutually-Refined Proposal Extraction}, which employs mask queries and visual features to mutually refine each other, leading to the extraction of more detailed mask proposals. These high-quality class-agnostic proposals can generalize effectively to unseen classes, thereby reducing the model's sensitivity to seen classes. 
Subsequently, we propose the \textbf{Generalization-Enhanced Proposal Classification}, an innovative approach that integrates generality constraints within the feature space, enhancing its generalizability to unseen classes. 
This includes integrating synthetic features to prevent over-specialization towards seen classes, and employing a multi-background prototype strategy for diverse background representations.
While these developments enhance zero-shot semantic segmentation by optimizing towards the zero-shot task objective, the exclusive use of seen class data in training inevitably leads to prediction bias. 
Therefore, we propose \textbf{Predictive Bias Correction} to filter potential unseen class proposals and adjust the corresponding prediction scores, thereby explicitly alleviating the prediction bias. 
Notably, our approach assists in judging potential unseen classes at the proposal level, first introducing a new effective step to the segmentation pipeline.

We evaluate AlignZeg in the Generalized Zero-Shot Semantic Segmentation (GZS3) setting and achieve an average improvement of 3.8\% in $\mathrm{hIoU}$, with a notable average increase of 7.1\% in $\mathrm{mIoU}(\mathcal{U})$ for unseen classes. 
Additionally, in the strict Zero-Shot Semantic Segmentation (ZS3) setting, which only considers unseen classes during testing, AlignZeg also achieves state-of-the-art (SOTA) performance in both $\mathrm{pAcc}$ and $\mathrm{mIoU}$. 
These experiments demonstrate that our model effectively mitigates the issue of objective misalignment. 

\section{Related Works}
\label{sec:related_works}

\noindent \textbf{Zero-shot visual recognition.}
Zero-shot visual recognition, aimed at identifying unseen categories, typically employs semantic descriptors like attributes~\cite{jayaraman2014zero} and word vectors~\cite{socher2013zero} to bridge seen and unseen categories. 
Classical methods~\cite{han2021contrastive, yue2021counterfactual, chen2021free, chen2021semantics, xu2022vgse, su2022distinguishing} create a shared space between features and semantic descriptors but often focus more on seen classes, leading to an \textbf{objective misalignment} issue. Recent works~\cite{huynh2020fine, xu2020attribute, liu2021goal, ge2022dual, chen2022transzero, liu2023progressive} address this by extracting attribute-related regions. 
However, the reliance on manually annotated attributes limits generalizability across diverse scenarios. 
Recently, large-scale vision-language models~\cite{jia2021scaling, li2019visualbert, su2019vl, radford2021learning} like CLIP~\cite{radford2021learning} have advanced by training on numerous image-text pairs.  
While CLIP's broad semantic coverage offers good zero-shot capabilities, its focus on image level limits its efficacy in pixel-level recognition.



\noindent \textbf{Zero-shot semantic segmentation.}
Zero-shot semantic segmentation~\cite{bucher2019zero, li2020consistent, zhang2021prototypical, he2023primitive, he2023semantic, zheng2021zero, liu2023delving, karazija2023diffusion, li2023tagclip, deng2023segment} is an emerging research task. Earlier methods~\cite{baek2021exploiting, cheng2021sign, gu2020context} focused on linking visual content with class descriptions by a shared space. 
Recently, the advent of CLIP~\cite{radford2021learning} has shifted focus to pixel-level zero-shot recognition using vision-language models. 
Recent methods like SimBaseline~\cite{xu2022simple} and Zegformer~\cite{ding2022decoupling} have introduced a two-stage zero-shot segmentation approach: class-agnostic mask proposals extraction followed by their zero-shot classification. While effectively scaling down CLIP's capabilities to pixel level, challenges persist in integrating cropped images with CLIP. OVSeg~\cite{liang2023open} bridged this gap with mask prompt tuning. However, the two-stage process is inefficient due to multiple CLIP encoder iterations. 
Models like ZegCLIP~\cite{zhou2023zegclip}, DeOP~\cite{han2023open}, and SAN~\cite{xu2023side} overcome this by utilizing the CLIP encoder directly. They not only simplified the process but also increased computational efficiency by necessitating just one forward pass. 
However, these approaches primarily focus on adapting CLIP for pixel-level tasks with classification loss on seen classes. 
This objective misaligns with the intrinsic nature of zero-shot tasks and often results in overfitting. This \textbf{objective misalignment} issue is more pronounced in the context of segmentation tasks due to their stronger supervision. 

\noindent \textbf{Prediction bias.}
To reduce objective misalignment in zero-shot visual recognition, addressing the classifier's prediction bias towards seen classes is a pivotal way.  
Researchers have explored strategies like advanced domain detector design~\cite{atzmon2019adaptive}, entropy-based score adjustments~\cite{min2020domain}, distance-based gating networks~\cite{kwon2022gating}, and generated-based distance analysis~\cite{yue2021counterfactual}, to distinguish between seen and unseen categories. However, adapting these methods to pixel-level tasks remains challenging. \cite{zhang2021prototypical} extended these ideas to zero-shot segmentation with an unknown prototype for unknown mask extraction. However, its final performance is notably sensitive to the accuracy of the unknown mask extraction. 




\begin{figure*}[!t]
  \centering
  \includegraphics[width=0.95\linewidth]{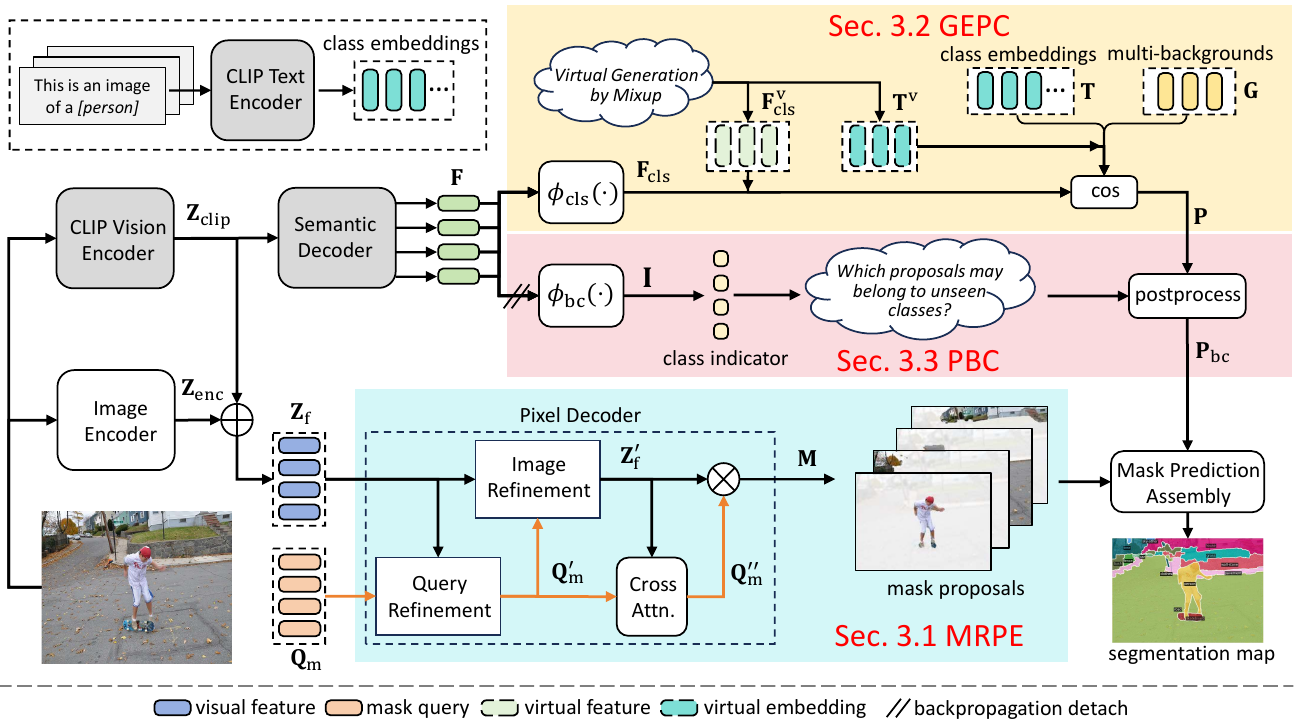}
   \caption{Overall framework. Our method mitigates the objective misalignment between semantic segmentation and zero-shot task through three main components: Mutually-Refined Proposal Extraction (MRPE), Generalization-Enhanced Proposal Classification (GEPC), and Proposal-based Bias Correction (PBC), the latter of which is applied during the inference process. Parameters in gray are fixed. }
   \label{fig:model}
\end{figure*}

\section{Method}
\noindent \textbf{Problem definition.}
Zero-shot semantic segmentation aims to extract transferable knowledge from seen classes $\mathcal{C}_\mathrm{seen}$ to subsequently recognize unseen classes $\mathcal{C}_\mathrm{unseen}$, where $\mathcal{C}_\mathrm{seen} \cap \mathcal{C}_\mathrm{unseen} = \varnothing$. To achieve this objective, during the training process, we leverage the CLIP model to derive semantic representations from class labels within $\mathcal{C}_\mathrm{seen}$. During the testing phase, we evaluate the model on both seen and unseen classes, \textit{i.e.}, $\mathcal{C}_\mathrm{test} = \mathcal{C}_\mathrm{unseen} \cup \mathcal{C}_\mathrm{seen}$. 

\noindent  \textbf{Overview.} 
Most existing methods prioritize improving recognition performance on seen classes, which deviates from the objectives of zero-shot tasks. In zero-shot semantic segmentation, this deviation, termed ``objective misalignment'', is exacerbated due to the strong supervision signals inherent in segmentation tasks. 
To realign the training process with zero-shot objectives, we introduce two key components: Mutually-Refined Proposal Extraction (MRPE) and Generalization-Enhanced Proposal Classification (GEPC), as detailed in \cref{sec_mask} and \cref{sec_cls}, respectively. MRPE refines mask proposals through interactions between queries and features, 
while GEPC enhances the generalizability of proposal features for each mask proposal, 
countering seen class dominance with virtual features and diverse backgrounds.
We also implement Predictive Bias Correction (PBC), outlined in \cref{sec_bias}, to explicitly reduce the model's bias towards seen classes by filtering potential unseen class proposals and adjusting their prediction scores.

\subsection{Mutually-Refined Proposal Extraction}
\label{sec_mask}
An intuitive idea to get the mask proposals is to use a simple transformer decoder like~\cite{cheng2021per, xu2022simple, ding2022decoupling} to decode masks from visual features via mask queries. However, mask queries of this technique lack sufficient interaction with visual features. 
This limitation can lead to challenges in reliably extracting class-agnostic masks, especially from samples that are complex and varied in nature. 
Therefore, we propose an enhanced strategy in the pixel decoder, where mask queries and visual features are mutually refined, which enables the extraction of high-quality, class-agnostic masks that are more effectively generalized to unseen classes.

Directly training a feature extraction network may over-fit the training set, while directly using CLIP features cannot guarantee good segmentation. 
Therefore, we extract complementary visual features $\mathbf{Z}_\mathrm{enc}$ and $\mathbf{Z}_\mathrm{clip}$ via a learnable image encoder and a fixed CLIP image encoder, respectively. Then we fuse them by $\mathbf{Z}_\mathrm{f} = \mathbf{Z}_\mathrm{enc} + \mathbf{Z}_\mathrm{clip} \in \mathbb{R}^{(\frac{H}{16}\times\frac{W}{16}) \times D}$ for following pixel decoder, where $H$ and $W$ are the input image's dimensions. 
We follow~\cite{zhou2023zegclip} to use prompt learning~\cite{jia2022visual} to fine-tune the CLIP features to further adapt CLIP to the current dataset. 

Then, we employ the queries to decode masks from the visual features. 
We first conduct query refinement shown in \cref{fig:model}. Specifically, we apply a cross-attention mechanism~\cite{carion2020end} to adjust the mask queries $\mathbf{Q}_\mathrm{m} \in \mathbb{R}^{N \times D}$ by
\begin{equation}\mathbf{Q}_\mathrm{m}^{\prime}=  \operatorname{softmax}\left(
\frac{\mathbf{Q}_\mathrm{m} \mathbf{W}_\mathrm{Q}\left(\mathbf{Z}_\mathrm{f} \mathbf{W}_\mathrm{K}\right)^{\top}}{\sqrt{D}}
\right) \mathbf{Z}_\mathrm{f} \mathbf{W}_\mathrm{V},
\label{eq.cross_attn}
\end{equation}
where $\mathbf{W}_\mathrm{Q}$, $\mathbf{W}_\mathrm{K}$, $\mathbf{W}_\mathrm{V}$ are the mapping functions for the queries, keys and values, and $D$ is their feature dimension. 
$\mathbf{Q}_\mathrm{m}$ is randomly initialized and is learnable. $N$ denotes the total number of queries, which corresponds to the extraction of $N$ distinct mask proposals.
Then we use residual connections and an MLP layer to prompt the queries by $\mathbf{Q}_\mathrm{m}^{\prime} = \operatorname{MLP}(\mathbf{Q}_\mathrm{m} + \mathbf{Q}_\mathrm{m}^{\prime})$. 
We also conduct image refinement to adjust the visual information. This process first includes a cross-attention mechanism to adapt the visual features and get $\mathbf{Z}_\mathrm{f}^{\prime}$ via \cref{eq.cross_attn}.
The only difference is that we obtain queries from $\mathbf{Z}_\mathrm{f}$, and get keys and values from $\mathbf{Q}_\mathrm{m}^{\prime}$. Then we get prompted visual features by $\mathbf{Z}_\mathrm{f}^{\prime} = \operatorname{MLP}(\mathbf{Z}_\mathrm{f} + \mathbf{Z}_\mathrm{f}^{\prime})$.

Finally, we apply the mutually prompted mask queries and visual features to get final queries $\mathbf{Q}_\mathrm{m}^{\prime\prime}$, which can also refer to the process expressed by \cref{eq.cross_attn}. Then we generate final mask proposals by $\mathbf{M} = \mathbf{Q}_\mathrm{m}^{\prime\prime}\cdot\mathbf{Z}_\mathrm{f}^{\prime\top}$. 



\subsection{Generalization-Enhanced Proposal Classification}
\label{sec_cls}
\begin{figure}[t]
  \centering
  \includegraphics[width=0.85\linewidth]{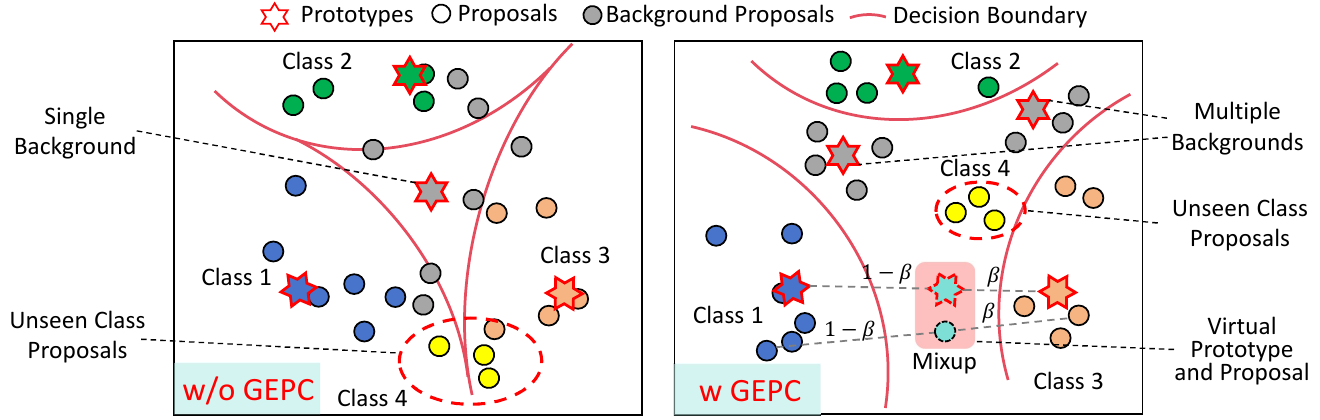}
   \caption{
   Without GEPC, driven mainly by classification loss, the model focuses on seen classes, allowing them to dominate the feature space. GEPC introduces synthetic features and diverse backgrounds, 
   helping provide a more generalizable feature space.   
   }
   \label{fig:mixup_back}
\end{figure}

After generating mask proposals, we extract the corresponding proposal features and use CLIP for zero-shot classification. However, traditional methods mainly use classification loss on seen classes, which does not align with the objectives of zero-shot tasks. 
This approach is likely to result in a wide-span occupancy of the feature space by seen classes, as illustrated in~\cref{fig:mixup_back} (left), ultimately impacting model generalization negatively. 
To address this, we introduce two straightforward yet effective strategies: \textbf{feature expansion strategy} and \textbf{background diversity strategy}. The former generates features outside the distribution of seen classes, while the latter focuses on learning diversified background prototypes. 
Collectively, these strategies assist in expanding the feature space available for unseen classes, thereby aiding in their better recognition. 

\noindent \textbf{Class embedding and proposal feature extraction.}
We first utilize the CLIP text encoder to extract class embeddings $\mathbf{T} \in \mathbb{R}^{C \times D}$ as the class prototypes, where $C$ denotes the number of classes. To account for categories beyond the predefined set, we concurrently learn a background token $\mathbf{g} \in \mathbb{R}^{1 \times D}$.
To maximize the generalization capacity of CLIP's visual information, we follow the approach in~\cite{xu2023side} to reuse the shadow [CLS] token of the CLIP image encoder as the prediction queries and employ the last layers of the CLIP image encoder as the semantic decoder. 
More detailed descriptions can be found in~\cite{xu2023side}. The output proposal features are defined as $\mathbf{F} \in \mathbb{R}^{N \times D}$. Subsequently, we utilize a feed-forward network $\phi_\mathrm{cls}(\cdot)$ to refine the features and obtain $\mathbf{F}_\mathrm{cls} = \phi_\mathrm{cls}(\mathbf{F}) \in \mathbb{R}^{N \times D}$.
Thus, the prediction scores can be derived as 
$\mathbf{P} = [\mathbf{F}_\mathrm{cls}\cdot\mathbf{T}^\top, \mathbf{F}_\mathrm{cls}\cdot\mathbf{g}^\top] \in \mathbb{R}^{N \times (C+1)}$.

\noindent \textbf{Feature expansion strategy.}
Our objective is to enhance the feature space originating from seen classes by synthesizing features beyond those classes. 
Mixup techniques like vanilla mixup~\cite{zhang2017mixup} and manifold mixup~\cite{verma2019manifold} provide valuable insights. However, vanilla mixup's image-level interpolation falls short for our segmentation model, whereas manifold mixup's feature-level interpolation better meets our requirements.
Moreover, diverging from the manifold mixup's strategy of blending one-hot class labels, we additionally mix class embeddings while interpolating visual features. This allows us to maintain alignment between visual features and class embeddings even after interpolation. 
Specifically, we first collect seen class proposal features $\mathbf{F}_\mathrm{cls}^\mathrm{b,s} \in \mathbb{R}^{N_\mathrm{b,s} \times D}$ and corresponding class embeddings $\mathbf{T}^\mathrm{b,s} \in \mathbb{R}^{N_\mathrm{b,s} \times D}$ from a mini-batch. Both sets are then shuffled to produce $\dot{\mathbf{F}}_\mathrm{cls}^\mathrm{b,s}$ and $\dot{\mathbf{T}}^\mathrm{b,s}$, respectively. Following this, we blend the original and shuffled features to synthesize the mixed features shown in~\cref{fig:mixup_back} (right):
\begin{equation}
\begin{aligned}
& \mathbf{F}_\mathrm{cls}^\mathrm{v}=\beta\cdot\mathbf{F}_\mathrm{cls}^\mathrm{b,s}+(1-\beta)\cdot\dot{\mathbf{F}}_\mathrm{cls}^\mathrm{b,s}, \\
& \mathbf{T}^\mathrm{v}=\beta\cdot\mathbf{T}^\mathrm{b,s}+(1-\beta)\cdot\dot{\mathbf{T}}^\mathrm{b,s},
\end{aligned}
\end{equation}
where $\beta \in [0,1]$ is sampled from a Beta distribution Beta($\alpha,\alpha$), $\mathbf{F}_\mathrm{cls}^\mathrm{v}$ and $\mathbf{T}^\mathrm{v}$ are the virtual proposal features and the corresponding virtual prototypes, respectively. 
We set $\alpha=3$ to encourage the virtual samples to deviate significantly from the seen classes.
To align the virtual features with virtual prototypes while separating them from seen classes, we calculate the logits by
\begin{equation}
\mathbf{P}^\mathrm{v}[i] = [\mathbf{F}_\mathrm{cls}^\mathrm{v}[i] \cdot \mathbf{T}^\top, 
\mathbf{F}_\mathrm{cls}^\mathrm{v}[i] \cdot \mathbf{T}^\mathrm{v}[i]^\top], i=0,1,\cdots, N_\mathrm{b,s},
\end{equation}
where $\mathbf{P}^\mathrm{v} \in \mathbb{R}^{N_\mathrm{b,s} \times (C+1)}$. The labels $\mathbf{Y}^\mathrm{v}$ of these generated features are all set to class $C+1$. Finally, we employ a cross-entropy loss $\mathcal{L}_\mathrm{vir}=\mathrm{CE}(\mathbf{P}^\mathrm{v},\mathbf{Y}^\mathrm{v})$ to optimize. By generating virtual features that lie outside the distribution of seen classes, this strategy tends to reserve more feature space for categories outside the seen class, thereby facilitating better generalization to unseen classes.

\noindent \textbf{Background diversity strategy.}
Another drawback of recent methods lies in the insufficient representation of complex and varied background categories using a single fixed prototype. To address this issue, we introduce multiple background prototypes to preserve the diversity of backgrounds.
Specifically, we set $M$ background prototypes $\mathbf{G} = [\mathbf{g}_1;\mathbf{g}_2;\cdots;\mathbf{g}_M] \in \mathbb{R}^{M \times D}$. For each proposal feature $\mathbf{f}_\mathrm{cls} \in \mathbb{R}^{1 \times D}$, the multiple background logits are calculated by $\mathbf{p}_\mathrm{mg} = \mathbf{f}_\mathrm{cls}\cdot\mathbf{G}^\top \in \mathbb{R}^{1 \times M}$. We then obtain the final background score by weighted summation by $p_\mathrm{g} = \operatorname{sum}(\operatorname{softmax}(\mathbf{p}_\mathrm{mg}) \odot \mathbf{p}_\mathrm{mg})$,
where $\odot$ is the element-wise product. 
Then the final logits are $\mathbf{P} = [\mathbf{F}_\mathrm{cls}\cdot\mathbf{T}^\top, \mathbf{p}_\mathrm{g}] \in \mathbb{R}^{N \times (C+1)}$. To maintain the diverse distribution of the backgrounds, we constrain their distances by 
\begin{equation}
\mathcal{L}_\mathrm{reg} = \frac{2}{M(M-1)} \sum_{i=1}^{M-1} \sum_{j=i+1}^{M}\mathbf{g}_i\cdot\mathbf{g}_j^{\top}.
\end{equation}

As shown in~\cref{fig:mixup_back} and~\cref{fig:tsne}, our approach tends to offer a more expansive feature space, enhancing the generalization ability for unseen classes. 
Additionally, the discernible space also allows for more thorough Predictive Bias Correction in \cref{sec_bias}, thereby further mitigating prediction biases. 

\subsection{Predictive Bias Correction}
\label{sec_bias}
\begin{figure}[t]
  \centering
  \includegraphics[width=0.7\linewidth]{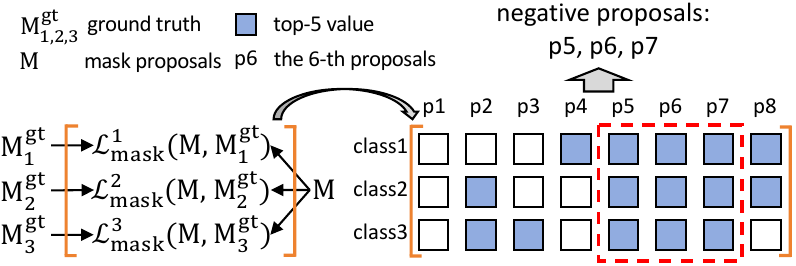}
   \caption{Example of negative proposal selection. In an image with 3 seen classes, p5, p6, and p7 are selected as negative proposals since they are in the top-K (K=5) based on loss for all seen classes contained in the image. 
   }
   \label{fig:bias}
\end{figure}

While the aforementioned methods optimize the model towards the objectives of zero-shot tasks, an inevitable prediction bias towards seen classes arises due to the model being trained on seen categories. 
To address this, we propose a simple yet effective \textbf{Predictive Bias Correction} to assist in identifying whether each proposal contains potentially unseen categories. 
Specifically, we employ $\phi_\mathrm{bc}(\cdot)$, a binary classification model comprising two fully connected layers followed by a sigmoid layer, to learn a class indicator for each proposal:
\begin{equation}
\mathbf{I} = \phi_\mathrm{bc}(\mathbf{F}) \in \mathbb{R}^{N \times 1},
\end{equation}
where each element of $\mathbf{I}$ lies in the range $[0,1]$. A value closer to 1 means that the seen class will not be included in the proposal. To train $\phi_\mathrm{bc}(\cdot)$, it is imperative to identify the positive and negative proposals and allocate the appropriate ground truth labels accordingly:
\begin{itemize}
    \item \textbf{Positive proposals.} 
    During training, proposals extracted by the model can be divided into seen class proposals and background proposals. We select the seen class proposals as the positive proposals and assign the label $0$ to them. 
    \item \textbf{Negative proposals.} 
    We select negative proposals from background proposals, particularly excluding the ones overlapping with seen class regions. 
    Specifically, we evaluate each proposal against seen classes present in the image by calculating a loss matrix $\mathcal{L}^{i}_\mathrm{mask}(\mathbf{M}, \mathbf{M}^\mathrm{gt}_i)$, comparing mask proposals $\mathbf{M}$ against the ground truth mask $\mathbf{M}^\mathrm{gt}_i$ for the $i$-th seen class in the image. A higher value in $\mathcal{L}^{i}_\mathrm{mask}$ suggests the proposal is less likely to contain the $i$-th seen class in the image. As shown in~\cref{fig:bias}, only proposals consistently ranking within the top-$K$ values across all $\mathcal{L}^{i}_\mathrm{mask}$ are considered as negative. This criterion ensures that our negative proposals are less likely to contain segments of any seen classes, thus more likely to represent pure background or undefined classes. We assign label 1 to these proposals. 
    \item \textbf{Others.} 
    Proposals that are neither selected as positive nor negative are excluded from optimization due to their low quality. 
\end{itemize}
After selecting the positive and negative proposals, we can optimize $\phi_\mathrm{bc}(\cdot)$ using a binary cross-entropy (BCE) loss:
\begin{equation}
\mathcal{L}_\mathrm{bc}=-\frac{1}{N_\mathrm{bc}} \sum_{i=1}^{N_\mathrm{bc}}{\left[y_i \log \left(I\right)+\left(1-y_i\right) \log \left(1-I\right)\right]},
\end{equation}
where $N_\mathrm{bc}$ is the number of selected proposals, and $I$ is the class indicator of the current proposal. 
In the inference phase, the trained $\phi_\mathrm{bc}(\cdot)$ can be used to obtain $\mathbf{I}$, enabling the use of a threshold $\gamma$ to identify possible unseen class proposals and suppress their seen class scores: 
\begin{equation}
\mathbf{P}_{\mathrm{bc}}[i,j] = 
\begin{cases} 
v_\mathrm{min}, & \text{if } \mathbf{I}[i] > \gamma \text{ and } \mathcal{C}_\mathrm{test}[j] \in \mathcal{S}, \\
\mathbf{P}[i,j], & \text{otherwise},
\end{cases}
\end{equation}
where $\mathcal{S}$ is the set of seen classes, $v_\mathrm{min}$ is the minimum value in $\mathbf{P}$, and $i$ represents the $i$-th proposal. 
$\mathbf{I}[i] > \gamma$ selects the potential proposals containing unseen classes, and $\mathcal{C}_\mathrm{test}[j] \in \mathcal{S}$ selects the seen class scores that need to be suppressed.
In this way, possible unseen class proposals can be well screened and their prediction bias for seen classes can be well mitigated.

\subsection{Optimization}
We follow Mask2former~\cite{cheng2022masked} and optimize our model using a classification loss $\mathcal{L}_\mathrm{ce}$ and a mask loss $\mathcal{L}_\mathrm{mask}$, which includes a dice loss and a binary cross-entropy loss. The overall loss is 
\begin{equation}
\mathcal{L} = \mathcal{L}_\mathrm{bc} + \lambda_1\cdot\mathcal{L}_\mathrm{ce} +  \lambda_2\cdot\mathcal{L}_\mathrm{mask} + \lambda_3\cdot\mathcal{L}_\mathrm{vir} + \lambda_4\cdot\mathcal{L}_\mathrm{reg}, 
\end{equation}
where $\lambda_1$, $\lambda_2$, $\lambda_3$ and $\lambda_4$ are coefficients.

\section{Experiments}
\subsection{Datasets and Evaluation Metrics}
\noindent \textbf{Datasets.}
We evaluate our method on three widely used datasets: \textbf{PASCAL VOC 2012}~\cite{everingham2012pascal}, \textbf{COCO-Stuff 164K}~\cite{caesar2018coco} and \textbf{PASCAL Context}~\cite{mottaghi2014role}. PASCAL VOC 2012 includes 10,582 images of 15 seen classes for training, and 1,449 images of 15 seen classes and 5 unseen classes for testing. COCO-Stuff 164K consists of 118,287 training images and 5,000 testing images. The classes are split into 156 seen and 15 unseen classes. PASCAL Context contains 59 semantic classes, of which 49 are seen and 10 are unseen. Its training and test sets contain 4,996 and 5,104 images, respectively.
Since the background category has no explicit semantics, they are ignored.
We also evaluate the generalization ability to other datasets. Following~\cite{xu2022simple,xu2023side,liang2023open}, we train the model on COCO-Stuff 164K with all 171 classes, and evaluate on five datasets: \textbf{A-150}~\cite{zhou2017scene}, \textbf{A-847}~\cite{zhou2017scene}, \textbf{VOC}~\cite{everingham2012pascal}, \textbf{P-59}~\cite{mottaghi2014role} and \textbf{P-459}~\cite{mottaghi2014role}. A-150 has 2K test images of 150 classes, and A-847 has the same images but is labeled with 847 classes. VOC indicates PASCAL VOC 2012, and has 1449 images of 20 classes. P-59 and P-459 are both from PASCAL Context, but contain 59 and 459 categories, respectively. 

\noindent \textbf{Evaluation metrics.}
Following~\cite{ding2022decoupling,xu2022simple,zhou2023zegclip}, we calculate the mean of class-wise intersection over union on both seen and unseen categories, resulting in $\mathrm{mIoU}(\mathcal{S})$ and $\mathrm{mIoU}(\mathcal{U})$, respectively. Simultaneously, we compute their harmonic mean, denoted as $\mathrm{hIoU}=\frac{2 \times \mathrm{mIoU}(\mathcal{U}) \times \mathrm{mIoU}(\mathcal{S})}{\mathrm{mIoU}(\mathcal{U})+\mathrm{mIoU}(\mathcal{S})}$, as a comprehensive evaluation metric. We also measure pixel-wise classification accuracy ($\mathrm{pAcc}$) for reference assessment.

\begin{table*}[t]
\centering
\caption{GZS3 results on three benchmarks. The highest scores are emphasized in bold.
$\mathrm{m}(\mathcal{U,S})$ represents $\mathrm{mIoU}(\mathcal{U,S})$, and $\mathrm{h}$ is the abbreviation of $\mathrm{hIoU}$.
} 
\setlength{\tabcolsep}{0.4mm}
{\scriptsize
\begin{tabular}{l|c|cccc|cccc|cccc}
\hline
\multicolumn{1}{c|}{\multirow{2}{*}{Methods}} & \multicolumn{1}{c|}{\multirow{2}{*}{Venue}}  & \multicolumn{4}{c|}{PASCAL VOC 2012} & \multicolumn{4}{c|}{COCO-Stuff 164K} & \multicolumn{4}{c}{PASCAL Context} \\ \cline{3-14} &
                         & $\mathrm{pAcc}$  & $\mathrm{m}(\mathcal{S})$  & $\mathrm{m}(\mathcal{U})$  & $\mathrm{h}$  & $\mathrm{pAcc}$   & $\mathrm{m}(\mathcal{S})$   & $\mathrm{m}(\mathcal{U})$  & $\mathrm{h}$  & $\mathrm{pAcc}$   & $\mathrm{m}(\mathcal{S})$  & $\mathrm{m}(\mathcal{U})$ & $\mathrm{h}$ \\ \hline
SPNet~\cite{xian2019semantic}    & CVPR'19          & -      & 78       & 15.6     & 26.1  & -      & 35.2     & 8.7      & 14    & -      & -        & -       & -    \\
ZS3~\cite{bucher2019zero}   & NeurIPS'19           & -      & 77.3     & 17.7     & 28.7  & -      & 34.7     & 9.5      & 15    & 52.8   & 20.8     & 12.7    & 15.8 \\
CaGNet~\cite{gu2020context} &  ACM MM'20                & 80.7   & 78.4     & 26.6     & 39.7  & 56.6   & 33.5     & 12.2     & 18.2  & -      & 24.1     & 18.5    & 21.2 \\
SIGN~\cite{cheng2021sign}  &  ICCV'21                  & -      & 75.4     & 28.9     & 41.7  & -      & 32.3     & 15.5     & 20.9  & -      & -        & -       & -    \\
Joint~\cite{baek2021exploiting} &  ICCV'21                   & -      & 77.7     & 32.5     & 45.9  & -      & -        & -        & -     & -      & 33       & 14.9    & 20.5 \\
ZegFormer~\cite{ding2022decoupling} &  CVPR'22             & -      & 86.4     & 63.6     & 73.3  & -      & 36.6     & 33.2     & 34.8  & -      & -        & -       & -    \\
SimBaseline~\cite{xu2022simple}  & ECCV'22         & 90     & 83.5     & 72.5     & 77.5  & 60.3   & 39.3     & 36.3     & 37.8  & -      & -        & -       & -    \\
SAN~\cite{xu2023side}  &     CVPR'23         & 94.7   & 92.3     & 78.9     & 85.1  & 58.4   & 40.1     & 39.2     & 39.6  & 81.5   & 53.5     & 56.6    & 55   \\
ZegCLIP~\cite{zhou2023zegclip} & CVPR'23               & 94.6   & 91.9     & 77.8     & 84.3  & 62     & 40.2     & 41.4     & 40.8  & 76.2   & 46       & 54.6    & 49.9 \\
DeOP~\cite{han2023open}  &  ICCV'23  & 92.5   & 88.2     & 74.6     & 80.8  & 62.2     & 38.0     & 38.4     & 38.2  & -   & -       & -    & - \\ 
MAFT~\cite{jiao2023learning}  & NeurIPS'23  & -   & 91.5     & 80.7     & 85.7  & -     & \textbf{40.6}     & 40.1     & 40.3  & -   & -       & -    & - \\ \hline
Ours   &    -                & \textbf{96.6}   & \textbf{93.9}     & \textbf{88.2}     & \textbf{91.0}    & \textbf{64.4}   & 40.2     & \textbf{50.0}       & \textbf{44.6}  & \textbf{82.2}  & \textbf{53.7}     & \textbf{61.7}    & \textbf{57.4} \\ \hline
\end{tabular}}
\label{tab:sota_gzsl}
\end{table*}

\subsection{Implementation Details}
Our image encoder is built upon an 8-layer Transformer architecture, with each layer having a dimension of 240 and a patch size of 16. Consistent with the model used in~\cite{zhou2023zegclip}, we employ the CLIP ViT-B/16 as our base model. 
Following~\cite{xu2023side}, the CLIP vision encoder and semantic decoder comprise the first 6 and last 3 layers of CLIP ViT-B/16, respectively.
Further details of the Semantic Decoder can be found in \textbf{the supplementary materials} and~\cite{xu2023side}. 
The input sizes for the CLIP Vision Encoder and Image Encoder are $320^2$ and $640^2$, respectively, to cater to their individual focuses on classification and segmentation. 
We set the batch size to 16. Our exploration is confined to inductive zero-shot learning, deliberately excluding the unseen pseudo labels generation and self-training process, as these do not align with real-world scenarios. The number of training iterations on PASCAL VOC, COCO-Stuff, and PASCAL Context datasets is 20K, 80K, and 40K, respectively. 
We follow~\cite{cheng2022masked} to set $\lambda_1=2$, $\lambda_2=5$. The parameter $\lambda_3$ and $\lambda_4$ are empirically set to $10^{-2}$ and $10^{-4}$, respectively. $K$ is set to 50. $N$ is set to 100, and $M$ is set to 20 for COCO-Stuff and 10 for others. 
Following~\cite{ding2022decoupling, xu2022simple, zhou2023zegclip}, the background score is not considered during the inference process. 

\subsection{Comparison with State-of-the-Art Methods}
\begin{figure}[t]
  \centering
  \includegraphics[width=0.7\linewidth]{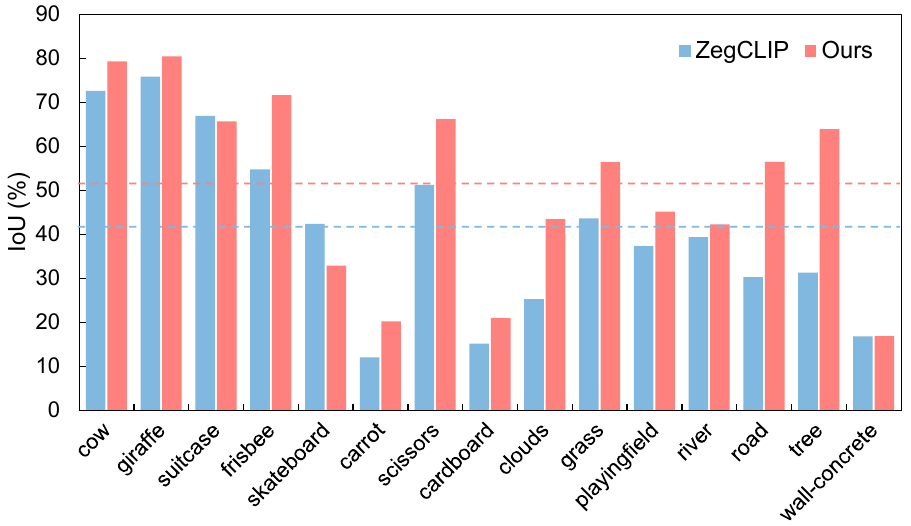}
   \caption{The $\mathrm{IoU}$ results for unseen classes on the COCO-Stuff. The dashed line represents the average result, \textit{i.e.}, $\mathrm{mIoU}(\mathcal{U})$.}
   \label{fig:coco_class}
\end{figure}

\noindent \textbf{Results of generalized zero-shot semantic segmentation.}
In \cref{tab:sota_gzsl}, we compare with prior approaches. 
Our method achieves superior performance across nearly all evaluated metrics, particularly in the pivotal $\mathrm{hIoU}$ metric where we surpass the SOTA performance on PASCAL VOC 2012, COCO-Stuff 164K, and PASCAL Context by margins of 5.3\%, 3.8\%, and 2.4\%, respectively. 
These gains are largely due to our improved recognition of unseen classes, as shown by our $\mathrm{mIoU}(\mathcal{U})$ metric outperformance by 7.5\%, 8.6\%, and 5.1\% on these datasets. 
This indicates that our design for proposal extraction, classification, and correction in the segmentation pipeline can effectively alleviate the objective misalignment issue. 
Moreover, our model also sets new benchmarks on the $\mathrm{pAcc}$, further demonstrating its exceptional segmentation capabilities. 
We also present results for each unseen class on the COCO dataset as shown in 
\cref{fig:coco_class}. Compared to the SOTA method, ZegCLIP, our approach shows improvements in almost all unseen categories. This also indicates that our method can effectively alleviate the objective misalignment issue.  

\begin{table}[t]
\centering
\begin{minipage}[b]{0.45\linewidth}
\centering
\caption{Results on ZS3 with unseen categories only. No prediction bias with this setting. }
\setlength{\tabcolsep}{0.8mm}
{\scriptsize
\begin{tabular}{l|cc|cc}
\hline
\multicolumn{1}{c|}{\multirow{2}{*}{Methods}} & \multicolumn{2}{c|}{VOC} & \multicolumn{2}{c}{COCO} \\ \cline{2-5} 
\multicolumn{1}{c|}{}                         & $\mathrm{mIoU}$          & $\mathrm{pAcc}$         & $\mathrm{mIoU}$         & $\mathrm{pAcc}$         \\ \hline
ZegFormer~\cite{ding2022decoupling}                                    & 81.3              & 91.4             & 64.9             & 81.3             \\
SimBaseline~\cite{xu2022simple}                                        & 86.7              & 94.7             & 67.1             & 82.9             \\
SAN~\cite{xu2023side}                                          & 94.1              & 97.8             & 76.9             & 84.8             \\
ZegCLIP ~\cite{zhou2023zegclip}                              & 86.3              & 93.9             & 70.8             & 77.7             \\ \hline
Ours                                          & 94.6              & 97.9             & 82.9             & 91.2             \\ \hline
\end{tabular}}
\label{tab:sota_zsl}
\end{minipage}
\hspace{0.15cm} 
\begin{minipage}[b]{0.48\linewidth} 
\centering
\caption{Results of generalization ability to other datasets.}
\setlength{\tabcolsep}{0.4mm}
{\scriptsize
\begin{tabular}{l|c|c|c|c|c}
\hline
\multicolumn{1}{c|}{Methods}   & A-847               & P-459                & A-150               & P-59                 & VOC                  \\ \hline
ZegFormer~\cite{ding2022decoupling} & -                     & -                     & -                     & 36.1                  & 85.6                 \\
SimBaseline~\cite{xu2022simple}   & 7                     & 8.7                   & 20.5                  & 47.7                  & 88.4                 \\
OVSeg~\cite{liang2023open}    & 7.1                   & 11                    & 24.8                  & 53.3                  & 92.6                 \\
SAN~\cite{xu2023side}      & 10.1                  & 12.6                  & 27.5                  & 53.8                  & 94                   \\
ZegCLIP~\cite{zhou2023zegclip}  & -                     & -                     & -                     & 41.2                  & 93.6                 \\ \hline
Ours      & 10.4                  & 16.2                  & 28.1 & {54.3} & {94.7} \\ \hline
\end{tabular}}
\label{tab:sota_crossdata}
\end{minipage}
\end{table}

\noindent \textbf{Results of zero-shot semantic segmentation.}
We also extend our evaluations to the zero-shot semantic segmentation setting, \textit{i.e.}, $\mathcal{C}_\mathrm{test} = \mathcal{C}_\mathrm{unseen}$, which inherently avoids the bias of classifying unseen classes as seen by only considering unseen categories. 
The results of other methods, as presented in \cref{tab:sota_zsl}, were obtained using their provided codes. 
\cref{tab:sota_zsl} reveals that our model consistently achieves optimal performance on both metrics across the two datasets. Notably, on the COCO dataset, our model achieves a relative improvement of 12.1\% and 6.0\% in $\mathrm{mIoU}$ compared to ZegCLIP and SAN, respectively. This enhancement in performance underscores our model's ability to effectively generalize knowledge from seen to unseen categories. 

\noindent \textbf{Results of generalization ability to other datasets.}
We also assess our model's ability to generalize across various datasets, 
a challenge due to varying data distributions as highlighted in recent studies~\cite{xu2022simple, xu2023side, han2023open}. 
In this scenario, unseen classes are more prevalent, thus the problem of prediction bias—which our PBC is adept at alleviating—is less prominent. 
Despite this, \cref{tab:sota_crossdata} demonstrates that our method maintains commendable performance, specifically when compared with the GZS3 SOTA method, ZegCLIP, with improvements of 13.1\% and 1.1\% on the P-59 and VOC datasets, respectively. This indicates that our approach can uphold its effectiveness in generalizing to new datasets.

\begin{table}[t]
\small
\centering
\caption{The impact of different components. PBC represents Predictive Bias Correction. MRPE indicates Mutually-Refined Proposal Extraction. 
FES (Feature Expansion Strategy) and BDS (Background Diversity Strategy) are key components of the GEPC (Generalization-Enhanced Proposal Classification).}
{\scriptsize
\begin{tabular}{l|cccc|cccc}
\hline
\multicolumn{1}{c|}{\multirow{2}{*}{Methods}} & \multicolumn{4}{c|}{PASCAL VOC 2012}                & \multicolumn{4}{c}{COCO-Stuff 164K}                 \\ \cline{2-9} 
\multicolumn{1}{c|}{}                         & \multicolumn{1}{l}{$\mathrm{pAcc}$} & $\mathrm{mIoU}(\mathcal{S})$ & $\mathrm{mIoU}(\mathcal{U})$ & $\mathrm{hIoU}$ & \multicolumn{1}{l}{$\mathrm{pAcc}$} & $\mathrm{mIoU}(\mathcal{S})$ & $\mathrm{mIoU}(\mathcal{U})$ & $\mathrm{hIoU}$ \\ \hline
Baseline                             & 94.1                     & 91.8    & 73.1    & 81.4 & 57.4                     & 38.0      & 38.5    & 38.3 \\
+PBC                                           & 95.1                     & 92.9    & 75.4    & 83.2 & 60.5                     & 37.9    & 44.1    & 40.8 \\
+PBC+MRPE                                      & 95.5                     & 92.5    & 82.1    & 87.1 & 62.1                     & 40.2    & 43.0      & 41.6 \\
+PBC+MRPE+FES                                  & 96.3                     & 93.3    & 85.4    & 89.2 & 63.5                     & 40.1    & 46.7    & 43.2 \\
+PBC+MRPE+FES+BDS                              & 96.6                     & 93.9    & 88.2    & 91.0   & 64.4                     & 40.2    & 50.0      & 44.6 \\ \hline
\end{tabular}}
\label{tab:ablation}
\end{table}

\begin{figure}[t]
  \centering
  \includegraphics[width=0.8\linewidth]{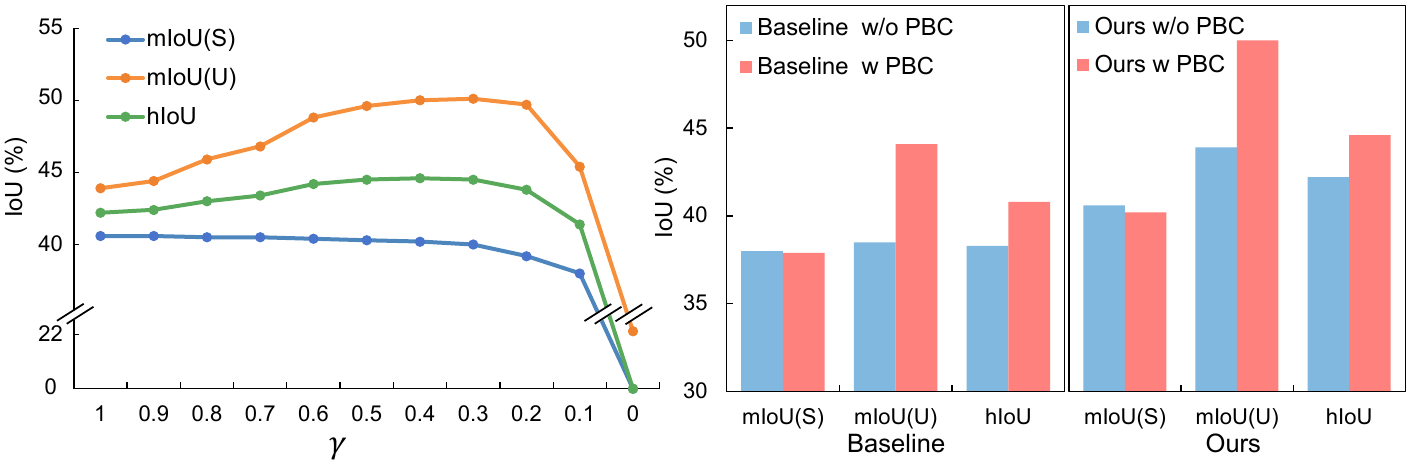}
   \caption{Left: Parameter $\gamma$ experiments in PBC. Right: Illustration of the effect of PBC.}
   \label{fig:bias_reduction}
\end{figure}

\begin{table}[t]
\centering
\begin{minipage}[b]{0.45\linewidth} 
\centering
\caption{Comparison with MaskFormer's pixel decoder. ``Ours w M-PDec'' indicates that we replace our pixel decoder with the pixel decoder from MaskFormer. }
{\scriptsize
\begin{tabular}{l|ccc}
\hline
\multicolumn{1}{c|}{Methods}       & $\mathrm{mIoU}(\mathcal{S})$ & $\mathrm{mIoU}(\mathcal{U})$ & $\mathrm{hIoU}$ \\ \hline \hline
Ours w M-PDec & 93.6 & 85.9   & 89.6 \\
Ours          & 93.9 & 88.2   & 91.0 \\ \hline
\end{tabular}}
\label{tab:mutual_eval}
\end{minipage}
\hspace{0.2cm} 
\begin{minipage}[b]{0.45\linewidth} 
\centering
\caption{Validation of the quality of class-agnostic mask proposals. We assign all the proposals with ground-truth class labels.}
{\scriptsize
\begin{tabular}{l|ccc}
\hline
Methods       & $\mathrm{mIoU}(\mathcal{S})$ & $\mathrm{mIoU}(\mathcal{U})$ & $\mathrm{hIoU}$ \\ \hline \hline
Baseline      & 95.0 & 89.4   & 92.1 \\
Ours          & 95.7 & 92.0   & 93.8 \\ \hline
\end{tabular}}
\label{tab:seg_eval}
\end{minipage}
\end{table}

\subsection{Ablative Studies}
\noindent \textbf{Effect of different components.}
Our ablation study results are in \cref{tab:ablation}.
We establish our baseline model based on the framework depicted in \cref{fig:model}, where the pixel decoder is implemented using a straightforward transformer decoder, akin to Maskformer, and is optimized solely with $\mathcal{L}_\mathrm{ce}$ and $\mathcal{L}_\mathrm{mask}$. The initial incorporation of Predictive Bias Correction (PBC) resulted in an average 2.15\% increase in the $\mathrm{hIoU}$ across two datasets. 
This shows PBC effectively reassigns proposals.
Subsequent integration of MRPE (Mutually-Refined Proposal Extraction) yielded more accurate class-agnostic masks, enhancing both $\mathrm{pAcc}$ and $\mathrm{hIoU}$.
\cref{tab:mutual_eval} further demonstrates this point.
Additionally, we implement Feature Expansion Strategy (FES) and Background Diversity Strategy (BDS) to impose semantic classification constraints from two different perspectives. FES, by introducing virtual features, effectively mitigates the feature space's excessive occupancy by seen classes, 
leading to an average increase of 1.1\% in $\mathrm{pAcc}$ and 1.85\% in $\mathrm{hIoU}$. 
BDS further enhances feature space generalization by promoting background class diversity during training, which ultimately propels the model to achieve its optimal results. 
Moreover, \cref{fig:bias_reduction} (right) shows that PBC further reduces prediction bias, even with the effective model ``Ours w/o PBC''. This suggests that our training optimizations, such as FES and BDS, also enhance the feature space for PBC, as seen in \cref{fig:mixup_back} (right).


\noindent \textbf{Evaluation of Class-Agnostic Mask Quality.}
We assign correct class labels to mask results to assess class-agnostic mask quality only, shown in~\cref{tab:seg_eval}. Key observations include: (1) The Baseline achieves high accuracy, indicating satisfactory existing mask performance; (2) Our model surpasses Baseline, validating MRPE’s effectiveness in enhancing class-agnostic mask extraction; (3) 
Referencing~\cref{tab:seg_eval} and~\cref{fig:coco_class}, we find classification to be the primary obstacle to improving zero-shot performance, seconded by mask quality, echoing findings from~\cite{liang2023open}. Our method has made advancements in both aspects.

\begin{figure}[t]
  \centering
  \includegraphics[width=0.95\linewidth]{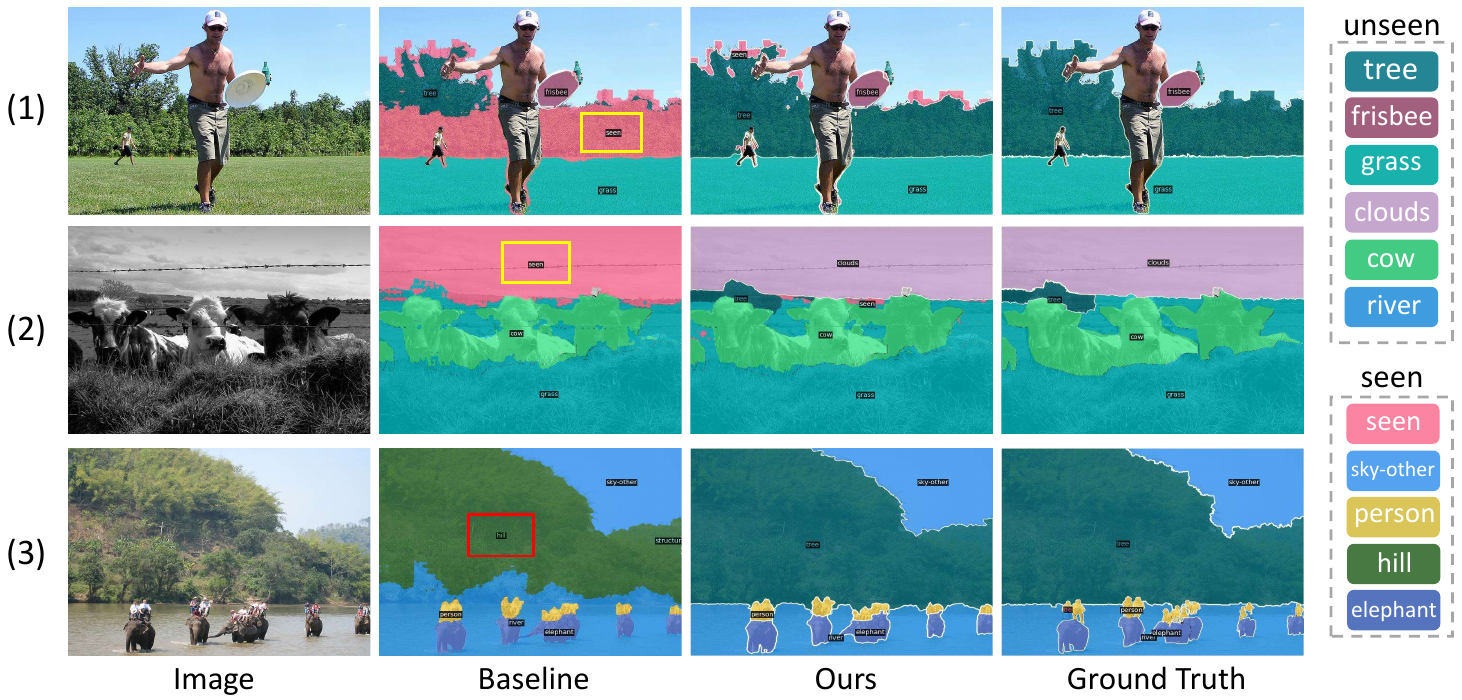}
   \caption{Visualization of segmentation on the COCO-Stuff 164K. Rows (1) and (2) depict the unseen classes segmentation results, 
   marking misclassified regions as ``seen'' label
   \begin{tikzpicture}
  \fill[color=seen_color] (0,0) rectangle (0.5,0.25); 
   \end{tikzpicture}
   . Row (3) illustrates the results for both seen and unseen categories. The four columns respectively display the original image, the segmentation results from the Baseline, the segmentation results from our method, and the ground truth.
   }
   \label{fig:seg_vis}
\end{figure}
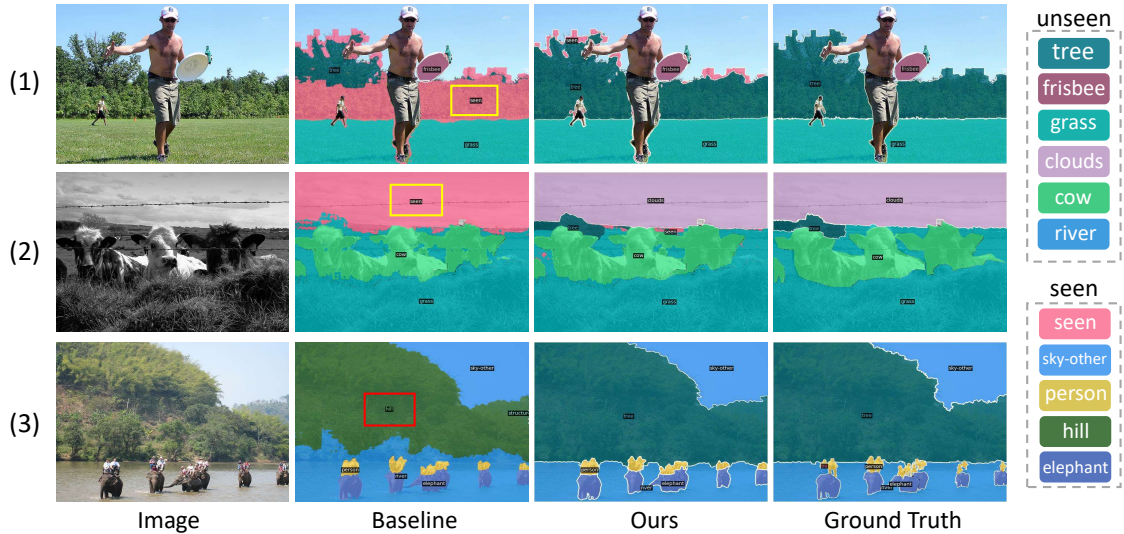

\noindent \textbf{Effect of $\gamma$.}
As shown in \cref{fig:bias_reduction} (left), as $\gamma$ decreases, an increasing number of proposals are categorized as unseen classes. This shift, while causing only minor fluctuations in $\mathrm{mIoU}(\mathcal{S})$, contributes to a substantial increase in $\mathrm{mIoU}(\mathcal{U})$, resulting in the optimal $\mathrm{hIoU}$ at $\gamma = 0.3$. This phenomenon indicates that our model can effectively uncover potential unseen class proposals, thereby aligning the inference objective more closely with the zero-shot learning goal. 

\noindent \textbf{Visualization.}
Segmentation results, shown in \cref{fig:seg_vis}, reveal model performance. 
In the first (1) and second (2) rows, the Baseline tends to erroneously classify unseen categories, such as trees and clouds, as the ``seen'' category. In contrast, our method can mitigate this bias, resulting in substantially improved results. The third row (3) highlights our method's superiority, showcasing its capability to generate more accurate segmentation masks (mask of the river) while concurrently making precise predictions for both seen and unseen categories. More visualization results can be found in \textbf{the supplementary materials}. 

\begin{figure}[t]
  \centering
  \includegraphics[width=1\linewidth]{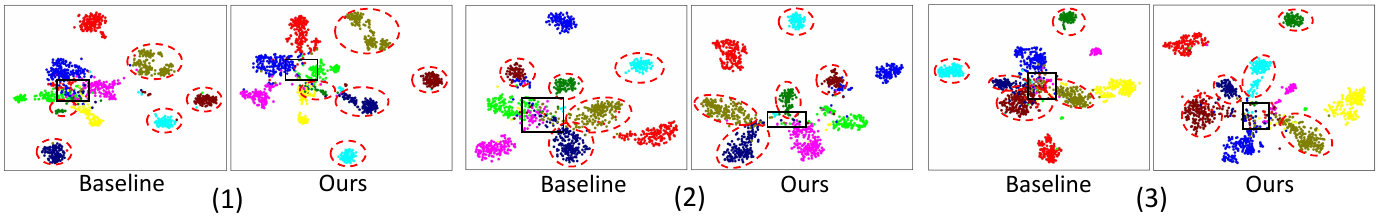}
   \caption{
    T-SNE~\cite{van2008visualizing} visualization of proposal features for five categories each from 156 seen and 15 unseen classes on COCO-Stuff 164K, presented in three random samplings (1)-(3) to mitigate the impact of outliers. Unseen categories are marked by red dashed boxes and feature entanglements are highlighted with black rectangles.
   }
   \label{fig:tsne}
\end{figure}

\noindent \textbf{T-SNE plot.} 
We visualize proposal features using our method versus the Baseline, as shown in~\cref{fig:tsne}. Our method's features display greater dispersion and clearer category margins, showing the effectiveness of Generalization-Enhanced Proposal Classification (GEPC) in enhancing feature distinctiveness. For example, \cref{fig:tsne} (3) illustrates that while the Baseline model's features for several unseen categories and one seen category are entangled, leading to misclassification, our approach effectively separates these features, improving distinguishability. 

\noindent \textbf{Limitations.}
Observations from \cref{fig:seg_vis} reveal that although our method effectively corrects the misclassified regions, the segmentation boundaries still exhibit some inaccurate areas (See the first and second rows of the third column) due to our ability to rectify only a subset of the proposals. This issue may be alleviated by incorporating post-processing techniques in segmentation to filter out such noise, which may lead to further enhancement.

\section{Conclusion}
In this paper, we introduce AlignZeg, an innovative architecture tackling the objective misalignment issue in zero-shot image segmentation by enhancing the entire segmentation pipeline. It integrates Mutually-Refined Proposal Extraction, Generalization-Enhanced Proposal Classification, and Predictive Bias Correction, enhancing unseen class segmentation while ensuring seen class performance. 
Notably, the Correction component aids in judging potential unseen classes at the proposal level, adding a novel step to the segmentation pipeline. 
Empirical results across various benchmarks demonstrate AlignZeg's superior performance in both Generalized and Strict Zero-Shot Semantic Segmentation. 
Our approach could also play a pivotal role in cross-modal pre-training, utilizing large-scale data for more effective zero-shot learning.



%
%
\bibliographystyle{splncs04}
\bibliography{main}
\end{document}


\title{Supplementary Material}


\author{}
\institute{}



\maketitle

In the supplementary material, we provide additional technical details in~\cref{sec:details}. \cref{sec:abla} presents ablation experiments on $\lambda_3$ and $M$. \cref{sec:vis} offers more visualization results, including filtered proposals of Predictive Bias Correction, more t-SNE~\cite{van2008visualizing} plots of proposal visual features, and segmentation results on two datasets. Finally, we give more discussions about relevant approaches in~\cref{sec:discussion}.

\section{More Technical Details}
\label{sec:details}
In our approach, we follow SAN~\cite{xu2023side} to design our Semantic Decoder, which utilizes the reuse of the [CLS] token and an attention bias $\mathbf{B}$ to guide the extraction of proposal features. We repurpose the [CLS] token $N$ times to act as a query that extracts proposal features corresponding to $N$ mask proposals from the visual features. To ensure that each extracted proposal feature is aligned with the mask proposals, we generate the attention bias $\mathbf{B}$ simultaneously with the mask proposals. This bias is then fed into the self-attention mechanism of the Semantic Decoder to extract features corresponding to the masks, as illustrated by the following formula:
$$
\mathbf{F}^\mathrm{SD} =  \operatorname{softmax}\left(
\frac{\mathbf{Q}^\mathrm{SD}\left(\mathbf{K}^\mathrm{SD}\right)^{\top}+\mathbf{B}}{\sqrt{D}} 
\right)\mathbf{V}^\mathrm{SD}.
$$
The formula is exemplified for a single layer to demonstrate the concept. $\mathbf{Q}^\mathrm{SD} = \mathbf{Q}_{\mathrm{[CLS]}} \mathbf{W}_\mathrm{Q}^{\mathrm{SD}}$, $\mathbf{K}^\mathrm{SD} 
 = \mathbf{Z}_\mathrm{clip}^{\mathrm{SD}} \mathbf{W}_\mathrm{K}^{\mathrm{SD}}$, $\mathbf{V}^\mathrm{SD} = \mathbf{Z}_\mathrm{clip}^{\mathrm{SD}}  \mathbf{W}_\mathrm{V}^{\mathrm{SD}}$, where $\mathbf{Q}_{\mathrm{[CLS]}}$ and $\mathbf{Z}_\mathrm{clip}^{\mathrm{SD}}$ are the input queries and visual features, respectively. $\mathbf{B}$ denotes the attention bias, which is derived by $\mathbf{B} = \mathbf{Q}_\mathrm{m}^{\prime\prime}\left(\mathrm{MLP}(\mathbf{Z}_\mathrm{f}^{\prime})\right)^{\top}$. This is analogous to the formula for generating mask proposals $\mathbf{M}$ presented at the end of Section 3.1 in the main text. Other details can refer to~\cite{xu2023side}. To unlock the potential of CLIP, we follow the prompt learning strategy used in ZegCLIP. We adjust the CLIP image encoder with 20 prompts on the COCO dataset, without applying this technique to other datasets. This is due to the broader category range and intricate scenes of the COCO dataset, which necessitates additional prompts for learning. 
 The experiments are conducted on eight NVIDIA V100 GPUs. 

Similar to most models~\cite{ding2022decoupling, xu2022simple, xu2023side, han2023open}, AlignZeg falls within the region-wise segmentation pipeline, originally pioneered by Maskformer~\cite{cheng2021per} and Mask2former~\cite{cheng2022masked}. In this pipeline, the model first extracts multiple ($N$) mask proposals and predicts each individually before integrating the results to obtain the final outcome. During training, the model is optimized using a classification loss $\mathcal{L}_{ce}$ and a mask loss $\mathcal{L}_{mask}$. The mask loss comprises a binary cross-entropy loss $\mathcal{L}_{bin}$ and a dice loss $\mathcal{L}_{dice}$, following Mask2former. To match these $N$ proposals with the ground truth, a bipartite matching-based assignment is employed, categorizing the proposals into the foreground (seen classes) and background proposals. This categorization is essential for the optimization process to proceed effectively.


\begin{figure}[t]
    \centering
    \begin{minipage}[b]{0.49\linewidth}
           \includegraphics[width=1\linewidth]{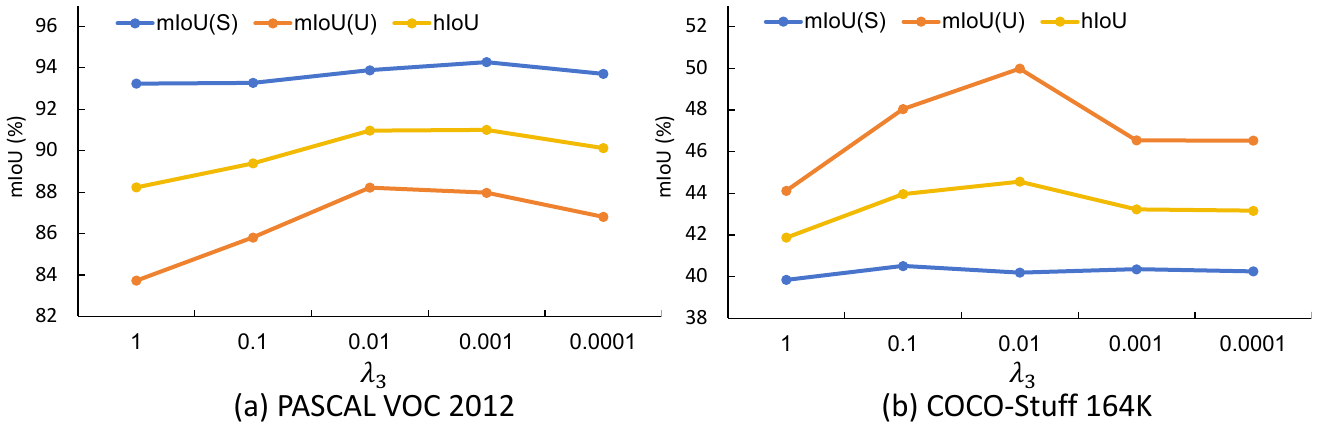}
           \caption{The results of different $\lambda_3$.}
           \label{fig:lambda}
    \end{minipage}
    \hfill 
    \begin{minipage}[b]{0.49\linewidth}
           \includegraphics[width=1\linewidth]{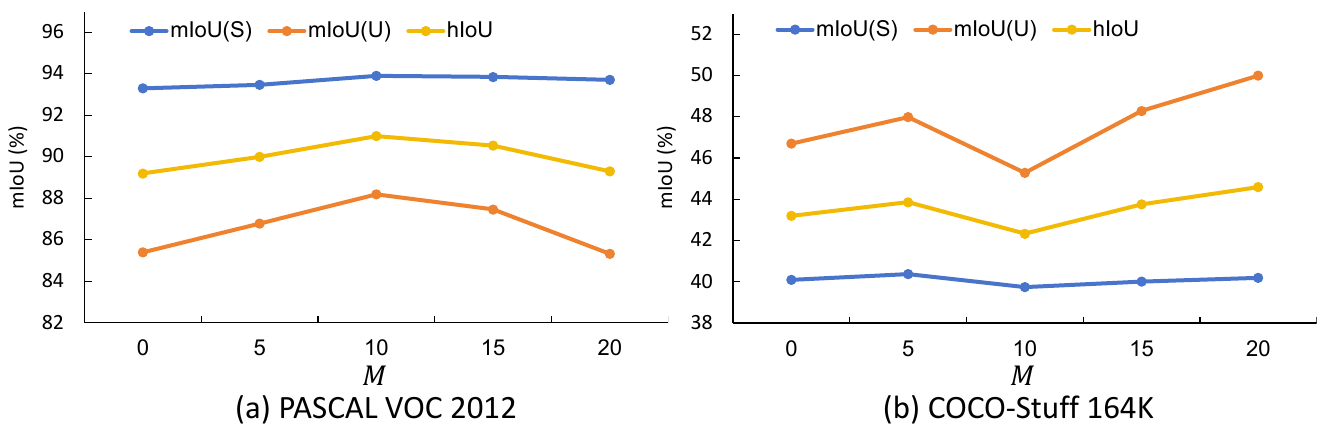}
           \caption{The results of different $M$.}
           \label{fig:m}
    \end{minipage}
\end{figure}

\section{More Ablation Experiments}
\label{sec:abla}
\subsection{Effect of $\lambda_3$}
\cref{fig:lambda} shows the impact of the weight $\lambda_3$ for the loss $\mathcal{L}_{vir}$ of feature expansion strategy. As depicted in the figure, a general trend is observed on both Pascal VOC 2012 and COCO-Stuff 164K datasets where the $\mathrm{hIoU}$ metric increases with the decrease of $\lambda_3$, reaching an optimal point before declining. Specifically, the optimal points for Pascal VOC 2012 are identified at $\lambda_3$ values of 0.01 and 0.001, while for COCO-Stuff 164K, the peak is at 0.01. Consequently, we set $\lambda_3 = 0.01$.

\begin{figure*}[!t]
  \centering
  \includegraphics[width=1\linewidth]{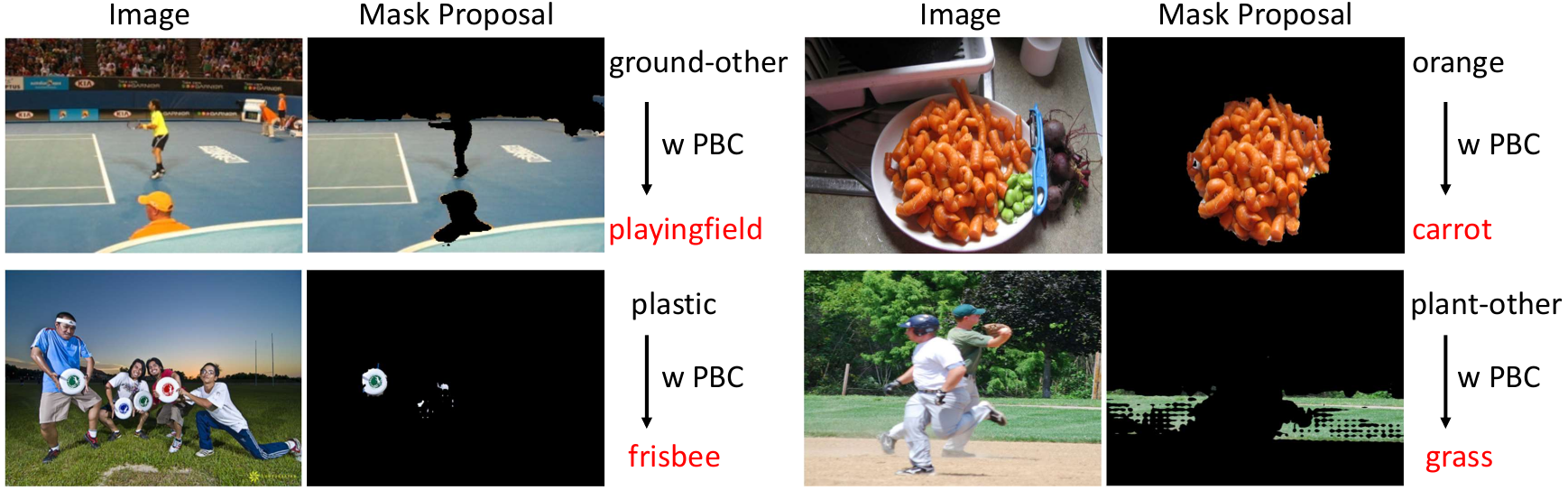}
   \caption{Examples corrected by Predictive Bias Correction (PBC). In the inference phase, PBC assists in filtering proposals that may include unseen classes, allowing for the recalibration of their prediction scores. The comparison illustrated by the arrows shows the transformation of proposal predictions before and after implementing PBC, with unseen classes highlighted in red.}
   \label{fig:pbc_testing}
\end{figure*}

\subsection{Effect of $M$}
$M$ represents the number of background prototypes used during the training process. As shown in~\cref{fig:m}, the optimal value of $M$ is 10 for PASCAL VOC 2012 and 20 for COCO-Stuff 164K. The smaller optimal value of $M$ for PASCAL VOC 2012 can be attributed to the relative simplicity of the dataset and the lesser variation in its backgrounds, implying that a smaller number of prototypes is sufficient to capture the diversity of the backgrounds.


\begin{figure*}[t]
  \centering
  \includegraphics[width=1\linewidth]{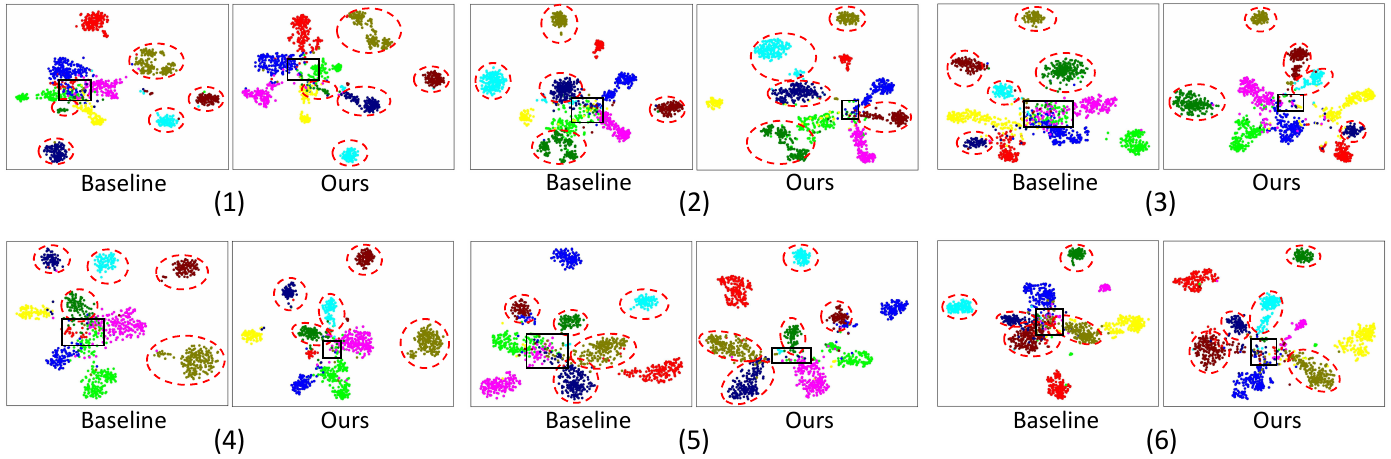}
   \caption{T-SNE~\cite{van2008visualizing} visualization of proposal features on COCO-Stuff 164K. Representative proposal features are selected through bipartite matching~\cite{cheng2021per}. In each plot, five categories are sampled from both 156 seen and 15 unseen classes, visualizing a total of 10 categories. To mitigate the impact of outliers, six random samplings (1)-(6) are presented. 
   Red dashed boxes mark unseen categories, while areas of feature entanglement are highlighted with black rectangles, indicating regions of interest.
   }
   \label{fig:tsne}
\end{figure*}

\section{More Visualizations}
\label{sec:vis}


\subsection{Effect of PBC}
The purpose of Predictive Bias Correction (PBC) is to train a binary classifier, enabling the filtering of proposals that potentially contain unseen classes during the inference phase, followed by score adjustments to reduce prediction bias. As demonstrated in~\cref{fig:pbc_testing}, Predictive Bias Correction (PBC) has the capability to correctly reclassify proposals that were mistakenly identified as seen classes into their accurate unseen class categories. For instance, it adjusts misclassifications from ``\textit{orange}'' to ``\textit{\textcolor{red}{carrot}}'' and from ``\textit{plastic}'' to ``\textit{\textcolor{red}{frisbee}}''. This correction process mitigates prediction bias and enhances the overall performance of the model.

\subsection{More Visualization Results of Visual Features}
We visualize the feature representations of proposals from both the Baseline and our method. As depicted in~\cref{fig:tsne}, the features from our method are generally more dispersed compared to the Baseline, exhibiting larger margins between categories. This demonstrates that our approach, particularly with the aid of Generalization-Enhanced Proposal Classification (GEPC), enhances the distinctiveness of features across different categories, thereby improving the semantic segmentation of unseen classes. However, it is important to note that despite the increased discriminative nature of the features, some proposal features still show signs of misclassification. This might be due to our model's reliance on fixed category prototypes (\textit{i.e.}, the output of the CLIP Text Encoder), which limits our mitigation of the impact of certain category prototypes being too close together, thereby hindering further generalization. This may be mitigated by adapting the category prototypes, which is a focal point for our future work.

\subsection{Comparison of Results on PASCAL VOC 2012}
\begin{figure*}[!t]
  \centering
  \includegraphics[width=1\linewidth]{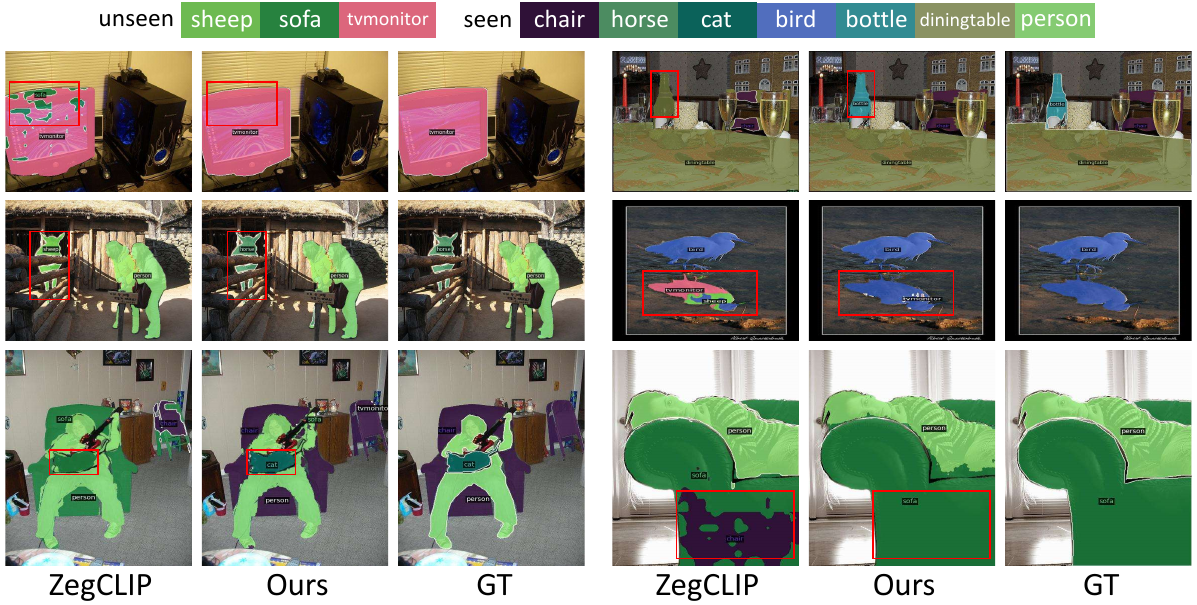}
   \caption{Visualization comparison on Pascal VOC 2012. Areas of interest are highlighted with red boxes. ``GT'' represents ground truth.}
   \label{fig:voc_compare}
\end{figure*}

We compared the segmentation results of ZegCLIP and our method on the PASCAL VOC 2012 dataset, as shown in~\cref{fig:voc_compare}. It is observable that ZegCLIP tends to misclassify some seen categories into semantically similar unseen categories, such as a horse being misclassified as a sheep in the second row. Additionally, its classification results are more susceptible to environmental influences, for instance, a cat being perceived as part of a sofa in the third row, and a bottle mistaken as part of a dining table in the first row. Conversely, our method effectively alleviates these issues. This improvement primarily stems from our enhancements across the entire segmentation pipeline, encompassing proposal extraction, classification, and correction phases.

\begin{figure*}[!t]
  \centering
  \includegraphics[width=1\linewidth]{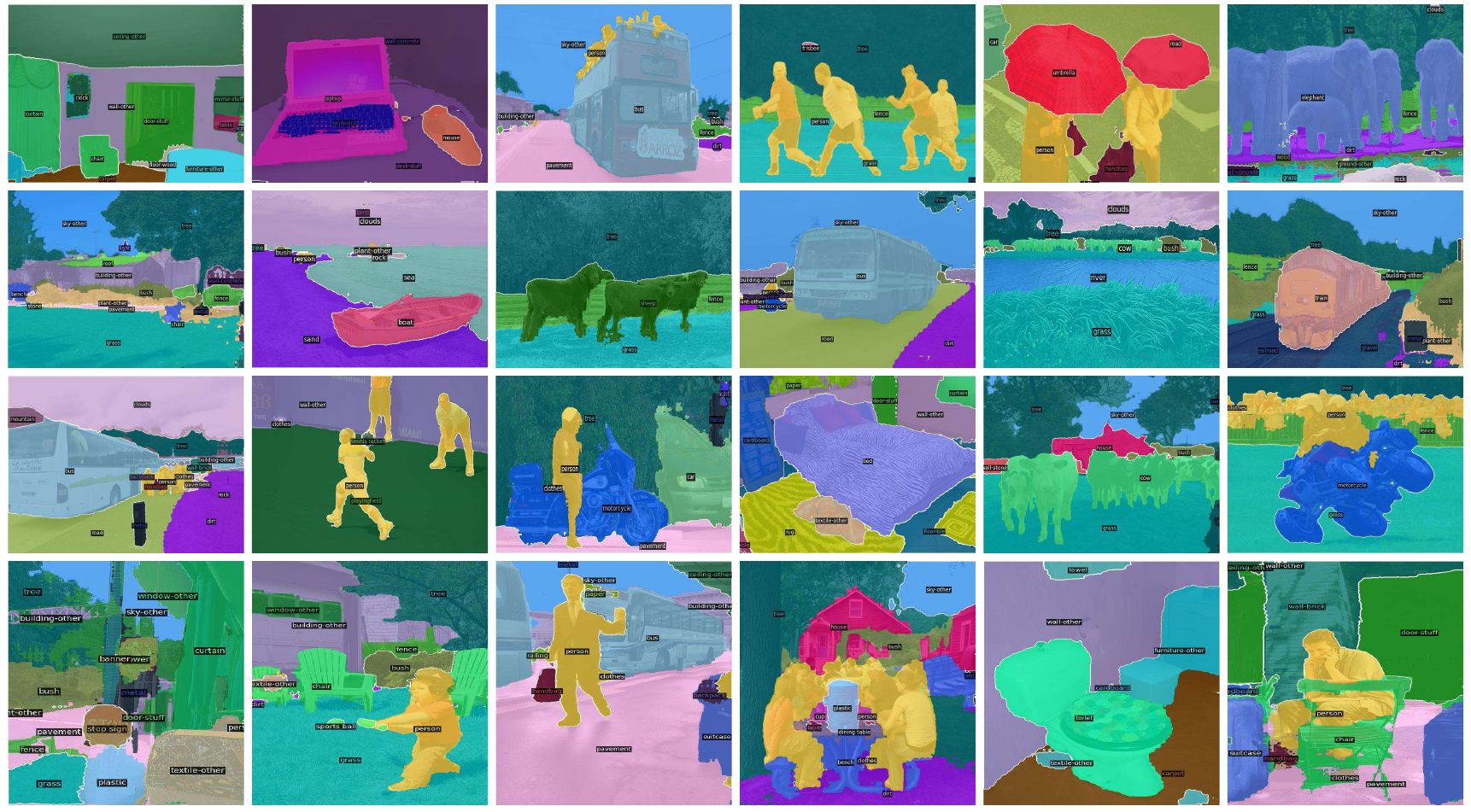}
   \caption{Visualization results on COCO-Stuff 164K. The outcomes demonstrate that AlignZeg is capable of adapting to a wide range of scenes, including indoor scenes, urban landscapes, and natural environments.}
   \label{fig:coco}
\end{figure*}

\subsection{More Visualization Results on COCO-Stuff 164K}
We show additional results of our method on the COCO-Stuff 164K dataset, demonstrating the effectiveness of AlignZeg in complex scenarios. As illustrated in~\cref{fig:coco}, our approach achieves commendable performance across diverse settings such as indoor scenes (\textit{e.g.}, first image in the first row and fourth image in the third row), urban landscapes (\textit{e.g.}, third image in the first row and first image in the third row), and natural environments (\textit{e.g.}, sixth image in the first row and first, second, third, and fifth images in the second row). This underscores the robust generalizability of our method, which is capable of reliably identifying both seen and unseen class regions in a variety of scenes. This success can be attributed to targeted improvements we make in the segmentation pipeline, including Mutually-Refined Proposal Extraction, Generalization-Enhanced Proposal Classification, and Predictive Bias Correction, allowing for accurate categorization on both seen and unseen classes across a broad range of scenarios. 

\section{More Discussions}
\label{sec:discussion}
We compare AlignZeg against multiple relevant methods, including ZegCLIP~\cite{zhou2023zegclip}, SAN~\cite{xu2023side}, DeOP~\cite{han2023open}, MAFT~\cite{jiao2023learning}, and PMOSR~\cite{zhang2021prototypical}. These comparisons are crucial for establishing the effectiveness and novelty of our approach.


We first examine ZegCLIP, a state-of-the-art method in Generalized Zero-shot Semantic Segmentation. The most critical among the designs of ZegCLIP is the Relationship Descriptor, which leverages the [CLS] token in CLIP's image encoder to preserve the inherent semantic information of CLIP. This indeed is an effective way to utilize CLIP's semantic capacity. Similarly, our Semantic Decoder, following the design of SAN~\cite{xu2023side}, also repurposes CLIP's [CLS] token. However, this alone is insufficient, because the optimization target still does not align with the zero-shot task's objectives, \textit{i.e.}, \textbf{the objective misalignment issue}. 
For instance, there is an absence of class-agnostic mask designs, a lack of feature-level optimization strategy despite focusing on classifying seen classes, and no targeted predictive bias correction during inference. 
Our approach addresses these by optimizing the segmentation pipeline holistically: promoting class-agnostic mask extraction through mutually refined mask queries and visual features, preventing overfitting to seen classes in the feature space by generating data and designing multiple backgrounds, and alleviating predictive biases by differentiating unseen classes during inference phase. 

Both SAN~\cite{xu2023side} and DeOP~\cite{han2023open} parallelly integrate the CLIP image encoder into their networks. SAN~\cite{xu2023side} innovatively harnesses CLIP's potential through [CLS] token reuse and mask bias techniques. Our approach follows SAN's methodology. DeOP incorporates the CLIP image encoder using generalized patch severance and classification anchor learning. However, both SAN and DeOP lack certain alignments with zero-shot task designs. For instance, (1) SAN's use of self-attention to adapt mask queries is easily dominated by the larger number of visual features. DeOP directly uses a transformer decoder, resulting in insufficient interaction between queries and features. (2) Both SAN and DeOP focus on optimizing proposals using a classification loss, which only considers correct categorization within seen classes. (3) Trained on seen classes, their inference inevitably biases towards these classes, a limitation neither SAN nor DeOP addresses. Our method improves upon these by extracting mask proposals via mutually enhanced mask queries and visual features, generalizing the feature space with generated data and multiple background prototypes, and screening potential unseen class proposals for targeted post-processing during inference, addressing the outlined issues more comprehensively.

MAFT~\cite{jiao2023learning}, a recently proposed plug-and-play approach, inherits SAN~\cite{xu2023side}'s strategy of reusing the [CLS] token to extract proposal features and using mask bias to guide the extraction process, effectively leveraging the zero-shot capabilities of the CLIP image encoder. Additionally, MAFT employs mask-aware loss and self-distillation loss to enhance the extraction of mask proposals and retain CLIP's performance. However, MAFT does not pay attention to the generalization strategy in the proposal features' feature space nor address prediction bias in the inference prediction stage, thus hindering its further improvement. 

One of our core contributions is the selection of unseen class proposals during the inference stage to mitigate prediction bias. While common in image-level tasks~\cite{atzmon2019adaptive, min2020domain, kwon2022gating, yue2021counterfactual}, this idea faces challenges in pixel-level tasks due to their complexity. PMOSR~\cite{zhang2021prototypical} made an initial attempt in zero-shot semantic segmentation by using an unknown prototype for unseen mask extraction, followed by semantic segmentation within this mask. However, this approach heavily relies on accurate unseen mask extraction. In contrast, following the region-wise segmentation pipeline of Maskformer~\cite{cheng2021per}, our method alleviates final prediction bias by filtering potential unseen class proposals from $N$ mask proposals. 
As our method directly processes individual proposal masks, the final results are less sensitive to the performance of unseen class proposal selection, making it more adaptable to a wider range of scenarios. 


\clearpage  

%
%
\bibliographystyle{splncs04}
\bibliography{main}